\documentclass[10pt,twocolumn,letterpaper]{article}

\usepackage{cvpr}
\usepackage{times}
\usepackage{epsfig}
\usepackage{graphicx}
\usepackage{amsmath}
\usepackage{amssymb}
\usepackage{multirow}

\usepackage{hyperref}
\cvprfinalcopy 

\def\cvprPaperID{6500} 

\begin{document}

\title{PCAN: 3D Attention Map Learning Using Contextual Information for Point Cloud Based Retrieval}

\author{Wenxiao Zhang, Chunxia Xiao\\
School of Computer Science, Wuhan University, China\\
{\tt\small wenxxiao.zhang@gmail.com cxxiao@whu.edu.cn}
}

\maketitle

\begin{abstract}
Point cloud based retrieval for place recognition is an emerging problem in vision field. The main challenge is how to find an efficient way to encode the local features into a discriminative global descriptor. In this paper, we propose a Point Contextual Attention
Network (PCAN), which can predict the significance of each local point feature based on point context. Our network makes it possible to pay more attention to the task-relevent features when aggregating local features. Experiments on various benchmark datasets show that the proposed network can provide outperformance than current state-of-the-art approaches.
\end{abstract}

\section{Introduction}



The task of visual localization is a core problem in computer vision, which can be widely applied in augmented reality\cite{pan2017simultaneous}, robot navigation \cite{mur2015orb,fu2018texture,yan2017distinguishing,yan2016geometrically}, and autonomous driving\cite{hee2013motion,liu2018robust}. Visual localization is always defined as follows: Given an image from a scene, the purpose is to predict the location where the photo is taken. 
The common process to complete this task is first to construct a large 2D image database, and then retrieval an image from the database which is closest to the query image. According to whether storing 3D point cloud in database, existing localization methods can be catagorized into image-retrieval based methods \cite{noh2017largescale,liu2018deep,radenovic2016cnn,muller2018geolocation} and direct 2D-3D matching based methods \cite{liu2017efficient,li2010location,toft2018semantic,toft2018semantic}. 
\begin{figure}[t]
\centering
\includegraphics[width=0.5\textwidth]{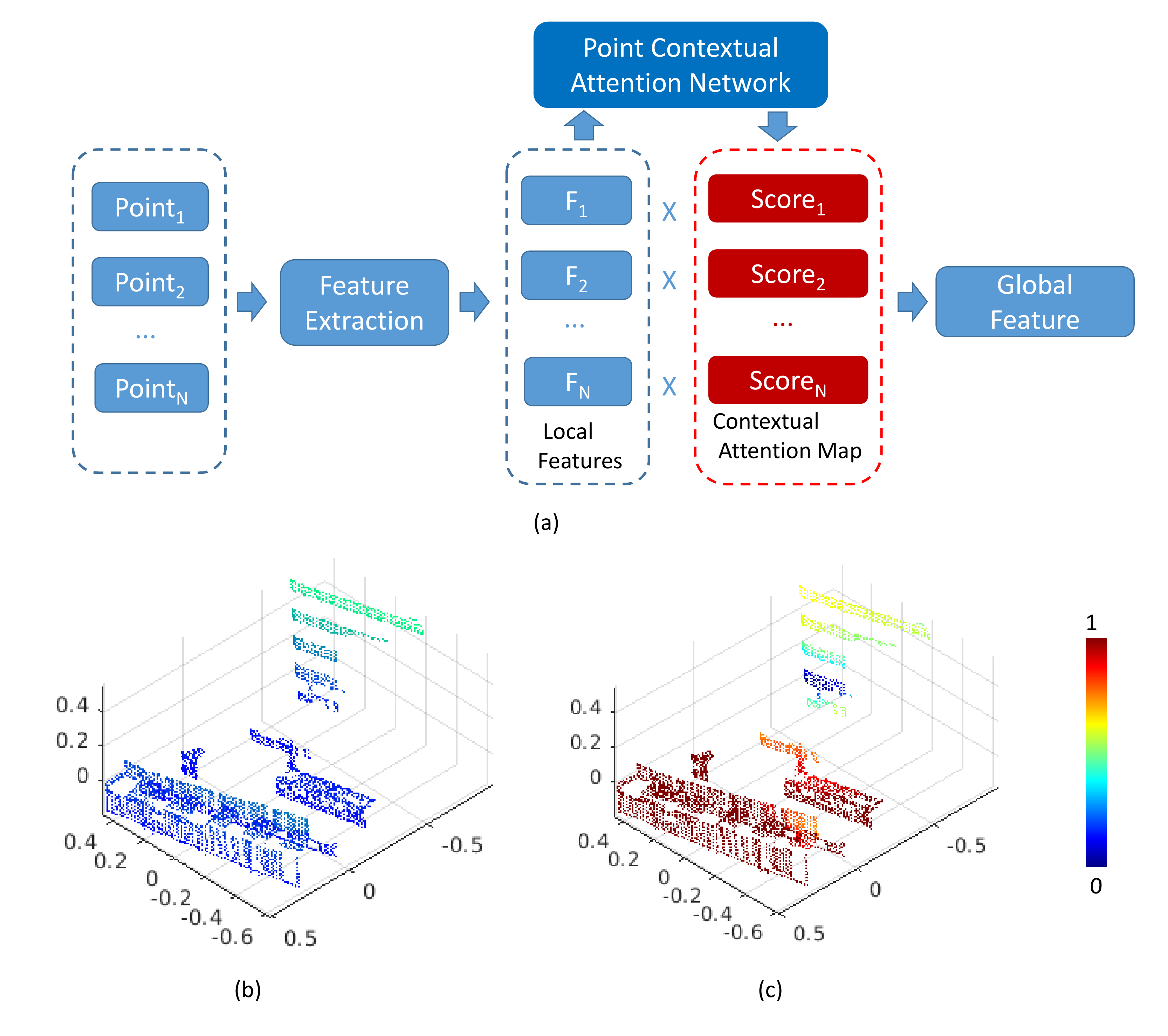}
\caption{(a) Our Point Contextual Attention Network (PCAN) takes the per-point local features and produces an attention map which estimates an attention score for each point based on the contextual information. (b) The input point cloud. (c) The visualization of the attention map.}
\label{Intro}
\end{figure}
All these methods are image based retrieval. However, image based retrieval always suffers from the illumination conditions. e.g. different day time, different climate environment, season changes. As the 3D scanned tech becomes mature, e.g. 3D Lidar sensor is widely used in robotics and autonomous driving, point cloud based retrieval for place recognition is first proposed in PointNetVLAD \cite{Uy_2018_CVPR}, a deep network for large-scale 3D point cloud retrieval. It uses PointNet \cite{qi2017pointnet} to extract local features which are then fed into a NetVLAD \cite{arandjelovic2016netvlad} to get the final discriminative global descriptor.

Actcually, when aggregating the local features into a global one, it is more reasonable to reweight the contributions for each local point feature before being aggregated to the global descriptor for distributing more attention to the task-relevant regions. For instance, in localization task, the time-varying objects such as pedestrians, or floating noisy pieces caused by unknown objects may lead to false retrieval results.

\begin{figure*}[t]
\centering
\includegraphics[width=1\textwidth]{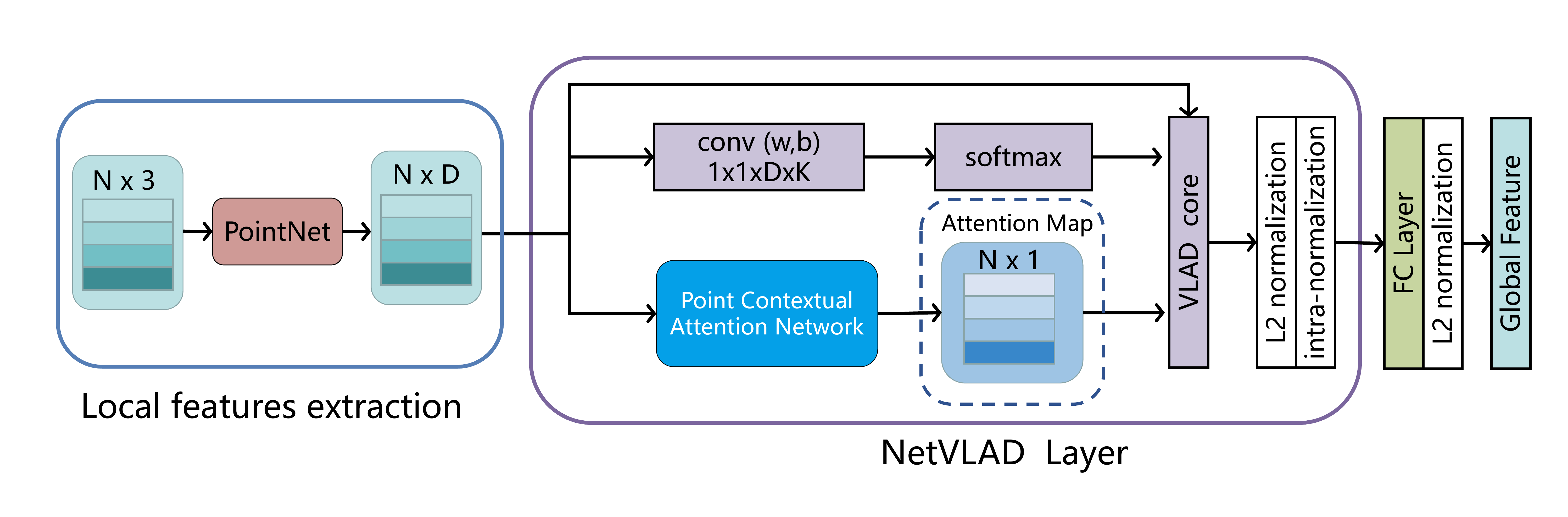}
\caption{Overall network architecture. The network takes $N$ points as input and uses the PointNet to extract local point features. Our Point Contextual Attention Network then takes the local point feaures and output a per-point attention map of size $N \times 1$. To efficiently leverage the attention map, we apply the attention map on the NetVLAD layer during feature aggregation. Finally, a fully connected(FC) layer is used to reduce the dimension of the feature vector and the L2 normalization is applied. }
\label{overview}
\end{figure*}

There has been a number of works focus on selecting the most important or interesting local features for localization in 2D image area \cite{jin2015predicting,knopp2010avoiding,schindler2007city}. Most of these works pay attention to the discriminative local features in feature space which can represent the image more efficiently. Kim et al.\cite{kim2017learned} addresses the problem that the significance of a local feature in a 2D image is largely effected by its context in the scene and proposes a Contextual Reweighting Network (CRN). 

We notice that context information is also essential in predicting the significance of local point features in point cloud based tasks. Since we do not use the color as the network input to avoid the false matching caused by different illumination conditions, each point more on their surroundings to infer point-wise semantic information. For instance, a sphere constructed by several points can be a street lamp when there is a lamp holder alike structure near by, while it also can be floating noises when it has no neighborhoods in a big search radius and is far away from all other points.

In this paper, we propose a novel context-aware reweighting network for 3D point cloud as illustrated in Fig. \ref{Intro}. The PCAN takes the local point features and produces an attention map which estimates a weight for each point based on the contextual information (Fig. \ref{Intro} (a)). Unlike 2D images, convolution operator can not be directly used in point clouds due to the sensitivity to the input permutation. Inspired by PointNet++ \cite{qi2017pointnet++}, we consider to use ball query search which depends on different query radius to aggregate muti-scale features. Both quality and quantity results on the benchmark datasets show the ability of our network in finding more important features to produce a more discriminative global descriptor.
\section{Related work}

\noindent{\bfseries Handcrafted 3D Descriptors. }Extracting robust local geometric descriptors has been a core problem in 3D vision. Some classic descriptors have been developed in the early years, such as Spin Images \cite{johnson1999using} and Geometry Histograms \cite{frome2004recognizing}. Recent works include Point Feature Histograms (PFH) \cite{rusu2008aligning}, Fast Point Feature Histograms (FPFH) \cite{rusu2009fast}, Signature of Histogram Orientations(SHOT) \cite{salti2014shot}. Some of these descriptors are already included in PCL \cite{rusu20113d}. Most of these hand-crafted descriptors are designed for specific tasks and they are sensitive to noisy, incomplete RGB-D images captured by sensors. Besides, these methods focus on extracting local descriptors which are not applicable to extracting global features due to the huge computation.

\noindent{\bfseries Learned 3D Global Descriptors. }With the breakthroughs of some learning based 2D vision tasks over the past few years, e.g. image classification, object detection, more and more researchers focus on representing 3D geometry using learning methods. In the early days, several works use volumetric representations as network input and develop learned descriptors for object retrieval and classification \cite{wu20153d,maturana2015voxnet,qi2016volumetric}. Recently researchers shifted to use raw point cloud \cite{yang2018foldingnet,su18splatnet,li2018sonet}. PointNet \cite{qi2017pointnet} directly handles point cloud and uses symmetric function to make the output invariant to the order permutation of the input points. PointNet++ \cite{qi2017pointnet++} leverages neighborhoods at multiple scales to capture local structures. Several network architectures on point cloud are proposed in succession mainly for classification and segmentation. \cite{wang2018deep} proposes Parametric Continuous Convolution, a new learnable operator that operates over non-grid structured data. All these works focus on extracting features from 3D data at a global level, but most of them aim at handling complete 3D models instead of 3D scanned data which is incomplete and noisy.

\noindent{\bfseries Learned 3D Local Descriptors. } 3DMatch \cite{zeng20173dmatch} uses voxel grid as input and introduced a 3D convolution network to distinguish the positive and negative pairs. Compact Geometric Features (CGF) \cite{khoury2017learning} uses histogram representation as network input to learn a compact local descriptor. PPFnet \cite{deng2018ppfnet} and PPF-FoldNet \cite{deng2018ppf} directly operate on points and uses a point pair feature encoding of the local 3D geometry into patches. 3DFeat-Net \cite{jian20183dfeat} proposes a weakly supervised network that learns both 3D feature detector and descriptor. However, these descriptors aiming at extracting local features are difficult to be applied in extracting global features due to the data increasing.

Point cloud based retrieval can be defined as a matching problem of the global descriptors of 3D point clouds.
Mikaela and Gim \cite{Uy_2018_CVPR} first propose PointNetVLAD which is a deep network combining PointNet \cite{qi2017pointnet} and NetVLAD \cite{arandjelovic2016netvlad} to extract the global descriptor from a scanned 3D point cloud for retrieval task. Though PointNetVLAD is more efficient to add a NetVLAD layer to the global feature than just using the vanilla PointNet architecture, it does not discriminate the local features which positively contribute to the final global feature representations. Based on these observations, we add a context-aware attention mechanism to the global feature extraction pipeline. 

\section{Network}
The overall architecture of our network is shown in Fig. \ref{overview}. We use the PointNet to extract local point features, and the input is a set of 3D points with coordinates which is denoted as  $P=\{p_1,...,p_N\}\in R^3$. Our Point Contextual Attention Network then takes the local point feaures and output a per-point attention map of size $N \times 1$. Instead of directly multiplying the local features by the corresponding scores on the attention map, we apply the attention map on the NetVLAD layer during feature aggregation. The details of our PCAN and feature aggregation process is described in Sec. ~\ref{CAM}.
\subsection{Local feature extraction}Similar to PointNetVLAD, we use PointNet as the basic network to extract per-point local features. The PointNet contains 2 transformation nets and 5 shared fully connected layers which is the same as the architecture used in PointNetVLAD.

\subsection{Attention map}

When computing an attention map for all the input points, we formulate the process as follows: After the local feature extraction, we get the per-point local features which are denoted as $F=\{f_1,...,f_N\}\in R^D$, where $D$ denotes the output feature dim of each point. Our goal is to learn a score matrix of size $N \times C$, where $C$ can be 1 which represents one score for one point, or the same as the local feature dim which represents one score for one channel. The attention score $s$ and the local feature $f$ at the space location $l$ is denoted as $s_l$ and $f_l$. We do a multiply between the local feature map and the attention map:
\begin{equation}
F_w=\{s_1 \times f_1,...,s_N \times f_N\}.
\end{equation}
$F_w$ denotes the reweighted local feature map. The final step is to aggregate the $F_w$ into a global descriptor.

To get an effective attention map, we proposed a point contextual network which concerns multi-scale context information to get an attention map. The output attention map is of size $N \times 1$ where we assign one score for one point.
\begin{figure}[t]
\centering
\includegraphics[width=0.48\textwidth]{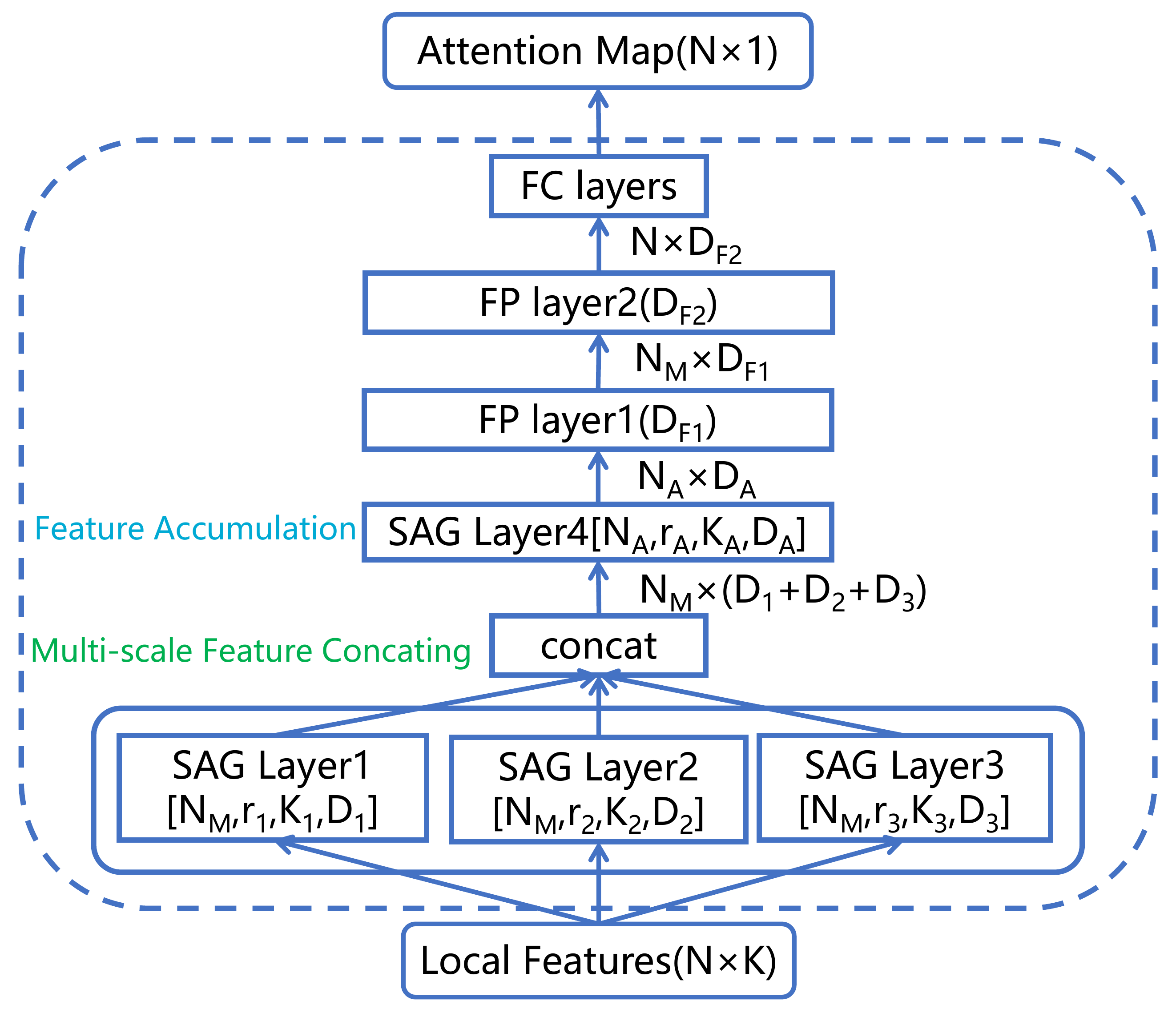}
\caption{PCAN architecture. PCAN takes the per-point local features and outputs an attention map representing the significance of spatial regions. It uses ball query search with different radius to aggregate multi-scale contextual information.}
\label{PCAN}
\end{figure}
\subsubsection{Contextual Attention Map}\label{CAM}

\noindent{\bfseries Multi-scale Contextual Information Aggregation } In CRN \cite{kim2017learned}, it uses context filters to obtain multi-scale contextual information implemented by convolution operator with different kernel sizes. However, convolution operator can not be directly applied on point cloud due to the orderless property of the point cloud. Inspired by PointNet++ \cite{qi2017pointnet++}, we use ball query search to aggregate multi-scale contextual information around a point. 

To group the contextual information of a point, it involves two main operations: sampling and grouping. We use the implementation of the sampling and grouping layer introduced in PointNet++. The sampling layer samples a sub set of $N'$ points from the origin $N$ points using iterative farthest point sampling (FPS) algorithm. The grouping layer uses ball query to find the neighborhoods. Given a radius $r$, ball query finds points that are within the radius $r$ to the query point. The input to this layer is a point set of size $N \times (d + C)$ and the coordinates of a set of sampled centroids of size $N' \times d$. The output are groups of point sets of size $N' \times K \times (d + C)$, where each group corresponds to a local region and $K$ is the number of points in the neighborhood of centroid points. Following the grouping layer, the PointNet layer with a pooling operation is used to aggregate the neighborhood information. So the final output data size is $N' \times (d + C')$. 
We also use the feature propagation(FP) layer introduced in PointNet++. It propagates features with distance based interpolation and across level skip links which can be considered as an upsampling operation. We use the FP layer to produce the final attention map. Refer to PointNet++ \cite{qi2017pointnet++} for more details.

For a clear expression, we use the following notations: SAG($N,r,K,D$) layer includes a sampling layer which samples $K$ local neighborhoods, a grouping layer with ball radius $r$ and a PointNet of with the final output dim $N \times D$. FP($D$) layer is the same with the FP layer introduced in PointNet++ and $D$ represents the final output feature dim.

The architecture of our PCAN is shown in Fig. \ref{PCAN}. To aggregate multi-scale features, we use three SAG layers for grouping the neighbor points with different ball query radius but with the same sampling number. 

We then concatenate these features to a multi-scale feature of size $N' \times D$. A following SAG layer groups all the points for feature accumulation. The accumulated feature is then fed to FP layers to recover the original point number. Finally, two fully connected layers are used to obtain the attention map of size $N \times 1$. Note that a sigmod activation function is applied to the attention map to restrict each attention score to $0-1$.

\noindent{\bfseries Feature Aggregation } Since our goal is to apply the attention map on the point cloud feature extraction pipeline, thus the final step is to aggregate the reweighted local features into a discriminative and compact global representation. 
PointNetVLAD uses a NetVLAD layer as a part of the network and it outperforms the original PointNet in retrieval task. Rather than directly multiplying the local features by the corresponding score on the attention map, we apply the attention map on the NetVLAD layer. The attention score $s$ and the local feature $f$ at the space location $l$ is denoted as $s_l$ and $f_l$. The NetVLAD layer learns K visual words $\{c_1,..,c_k|c_k \in R^D\}$, and the output subvector corresponding to the $c_k$ is denoted as $V_k$. $V_k$ is computed as the accumulation of differences between $f_l$ and $c_k$, weighted by a soft assignment $a_l^k$ which computes a weight of $f_l$ to $c_k$.
The original VLAD representation $V$ is denoted as:
\begin{equation}
V=[V_1,...,V_K],
\end{equation}
where
\begin{equation}
V_k=\sum_{j \in R} a_l^k(f_l - c_k).
\end{equation}
$R$ denotes a set of spatial location in the feature map.
We apply the learned attention map to the NetVLAD layer which reweights the VLAD representation $V_k$ as follows:
\begin{equation}
V_k=\sum_{l \in R}s_l \cdot a_l^k(f_l - c_k).
\end{equation}
The soft assignment operation is the same as the implementation described in PointNetVLAD. Our learned attention map can be seen as a reweighted pooling of 
the local feature. 
The final descriptor we got is of size $K \times D$, it is time-consuming in our task. Like PointNetVLAD, we use a fully connected layer to get a more compact global descriptor $V'$ with intra-nomalizing and then $L_2$ normalizing.
\begin{table}[t]
\centering
\begin{tabular}{|c|c|}
\hline
SAG layer1& $N=256, r=0.1, K=16, [16, 16, 32]$ \\
\hline
SAG layer2& $N=256, r=0.2, K=32, [32, 32, 64]$ \\
\hline
SAG layer3& $N=256, r=0.4, K=64, [32, 64, 64]$ \\
\hline
SAG layer4& $N=1, r=inf, K=256, [256, 512]$ \\
\hline
FP layer1& $[256, 128]$ \\
\hline
FP layer2& $[128, 128]$ \\
\hline
FC layers& $[1, 1]$ \\
\hline
\end{tabular}
\caption{Parameters of each layer in PCAN.}
\label{table_parameters}
\end{table}
\begin{figure*}[t]
\centering
\includegraphics[width=1\textwidth]{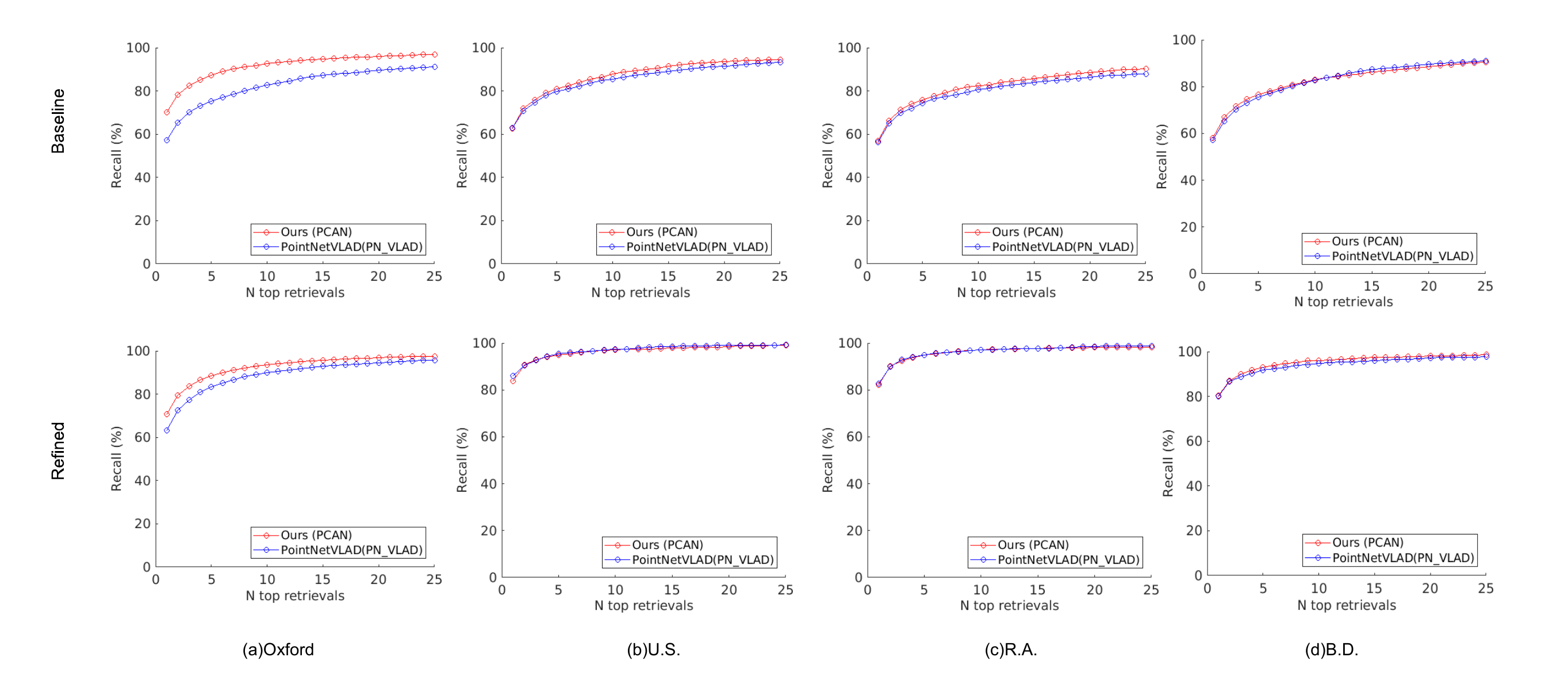}
\caption{Average recall of the networks.Top row shows the average recall of the baseline networks which are trained on Oxford. Bottom shows the average recall of the refined networks which are trained on Oxford, U.S and R.A.}
\label{recall_carve}
\end{figure*}
\section{Implements Details}
We show the parameters of each layer of our PCAN in Table \ref{table_parameters}. Note that the number sets $[l_1,..,l_d]$ in SAG layer and FP layer represent that it includes a succeeding PointNet layer of $d$ fully connected layers with width $l_i (i = 1,..., d)$. In Fig. \ref{PCAN}, we use $D$ to represent the width of the final fully connected layer for a clear illustration.

Since our method mainly focuses on the attention mechanism of the network, we use the same lazy quadruplet loss introduced in PointNetVLAD for a fair comparison. Most of the training settings of our network are the same with the settings in PointNetVLAD. Refer to the supplementary material for more details of the training settings.

\section{Experiments}
\subsection{Evaluation Datasets}
We use the benchmark datasets proposed in PointNetVLAD for evaluation. It builds a database of submaps for different areas and contains four datasets: Oxford RobotCar \cite{maddern20171} dataset and three in-house datasets, that is, a university sector (U.S.), a residential area (R.A.) and a business district (B.D.). They are all captured using a LiDAR sensor mounted on a car. PointNetVLAD removes the non-formative ground planes of all submaps and downsample them to fixed 4096 points. Each submap is tagged with an UTM coordinate. To get the ground-truth annotations for correct matches for each submap, point clouds are defined as positive pairs if they are at most 10m apart and negtive pairs if they are to be at least 50m apart. In evaluation process, the retrieved point cloud is regarded as a correct match if the distance is within 25m between the retrieved point cloud and the query point cloud.

Oxford Dataset consists of 21,711 submaps for training and 3030 submaps for testing from RobotCar dataset. The training submaps and testing submaps are splited at regular internvals of 10m and 20m.
The in-house datasets contain 400, 320, 200 submaps for training and 80, 75, 200 submaps for testing from U.S., R.A. and B.D. respectively. The training submaps and testing submaps are splited at regular intervals of 12.5m and 25m.

\begin{figure*}[t]
\centering
\includegraphics[width=1\textwidth]{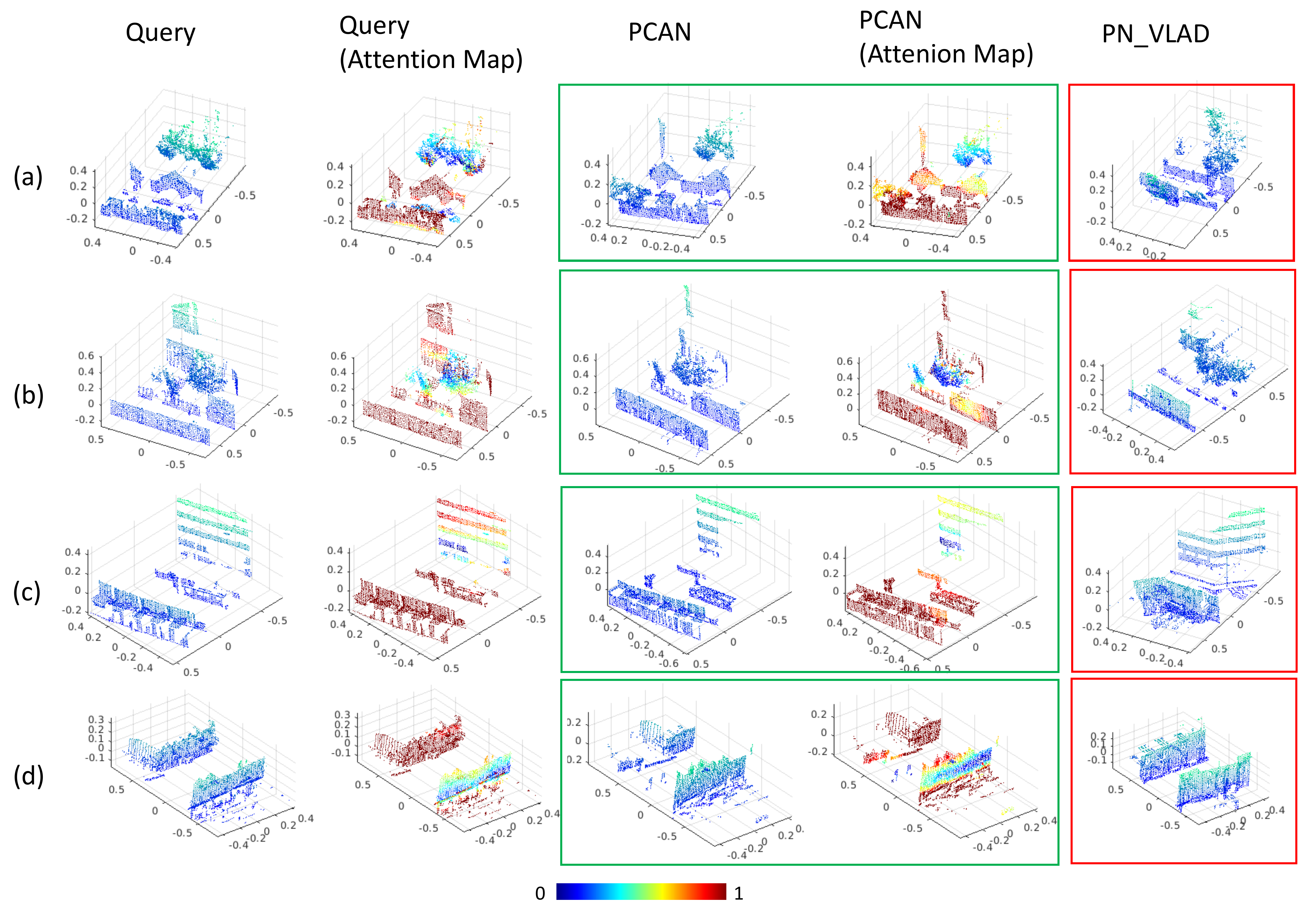}
\caption{Example retrieval results of our network on Oxford datasets. From left to right: query point cloud, attention map of the query point cloud, the top retrieved point cloud using PCAN, the top retrieved point cloud using PointNetVLAD. Green and red borders indicate correct and incorrect retrieved results, respectively.}
\label{compare}
\end{figure*}
\subsection{Results}

\noindent{\bfseries Baseline Network. } We first evaluate our attention network on the benchmark datasets proposed in PointNetVLAD. For a fair comparison with PointNetVLAD, the only difference between our network and PointNetVLAD is that we add the attention network into the NetVLAD layer, as illustrated in Fig. \ref{overview}. We also compare our network with other methods including the original PointNet architecture with the maxpooling layer (PN\_MAX) and the state-of-the-art PointNet trained for object classification on rigid objects in ModelNet \cite{wu20153d}(PN\_STD). The final output vectors of all the network are 256-dim. All above networks are trained using only the Oxford training dataset and tested on Oxford, U.S., R.A. and B.D. testing datasets. Table \ref{baseline} shows the top 1\% recall of each network on the bechmark datasets. 

Our baseline network significantly outperforms the other methods on all the datasets which achieves the best performance with an 83.81\% recall at top 1\%, exceeding the recall of
the PointNetVLAD by 2.8\%. It proves that our network can effectively explore the task-relevent local features on the Oxford datasets. Our network also outperforms the PointNetVLAD on the indoor-datasets with a margin of 1.2\%-1.5\%.
\begin{table}[t]
\centering
\begin{tabular}{|c|c|c|c|c|c|}
\hline
& PCAN &PN\_VLAD &PN\_MAX &PN\_STD \\
\hline
Oxford& \textbf{83.81} &81.01 & 73.44 &46.52\\
\hline
U.S.& \textbf{79.05} &77.83 & 64.64 &61.12 \\
\hline
R.A.& \textbf{71.18} &69.75 &51.92& 49.07 \\
\hline
B.D.& \textbf{66.82} &65.30 &54.74& 53.02 \\
\hline
\end{tabular}
\caption{ Baseline results showing the average recall (\%) at top 1\%
for each of the models.}
\label{baseline}
\end{table}
\begin{table}[t]
\centering
\begin{tabular}{|c|c|c|c|c|}
\hline
\multirow{2}*{ }&\multicolumn{2}{|c|}{Ave recall @1\%} &\multicolumn{2}{|c|}{Ave recall @1} \\
\cline{2-5}
&PCAN &PN\_VLAD & PCAN &PN\_VLAD \\
\hline
Oxford& \textbf{86.40} &80.70 & \textbf{70.72} & 63.33 \\
\hline
U.S.& 94.07 &\textbf{94.45} & 83.69 & \textbf{86.06} \\
\hline
R.A.& 92.27 &\textbf{93.07} & 82.26 & \textbf{82.65} \\
\hline
B.D.& \textbf{87.00} &86.48 & \textbf{80.31} & 80.11 \\
\hline
\end{tabular}
\caption{Refined network results showing the average recall (\%) at top 1\% and at top 1 after training on Oxford, U.S. and R.A.
for each of the models.}
\label{refined}
\end{table}
For PointNetVLAD, we use the results of the provided pre-trained model by the authors. For PN\_MAX and PN\_STD, we use the recall values reported in \cite{Uy_2018_CVPR}.

\begin{figure*}[t]
\centering
\includegraphics[width=1\textwidth]{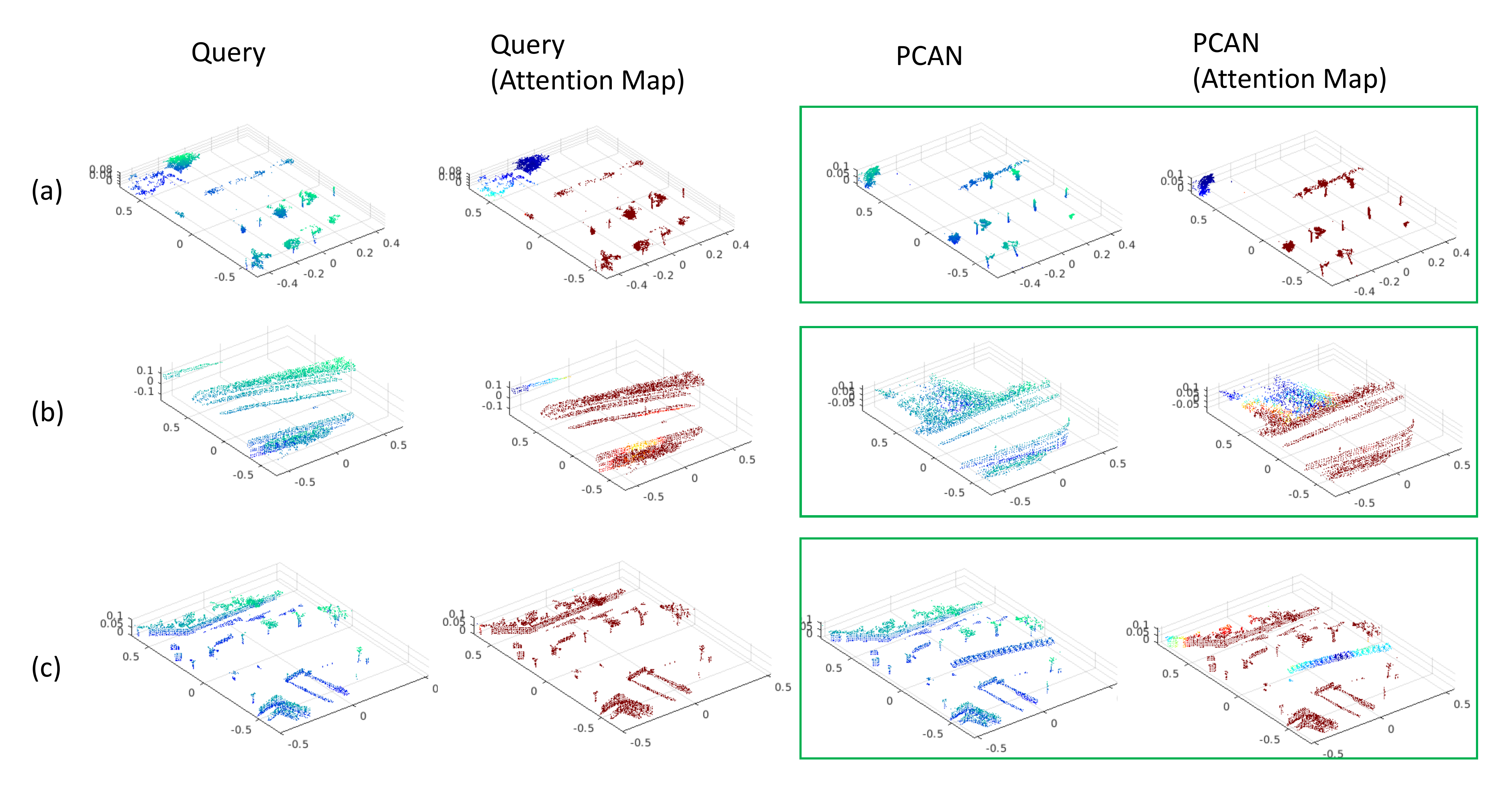}
\caption{Example retrieval results of our PCAN on in-house datasets.}
\label{good}
\end{figure*}
\begin{figure*}[t]
\centering
\includegraphics[width=1\textwidth]{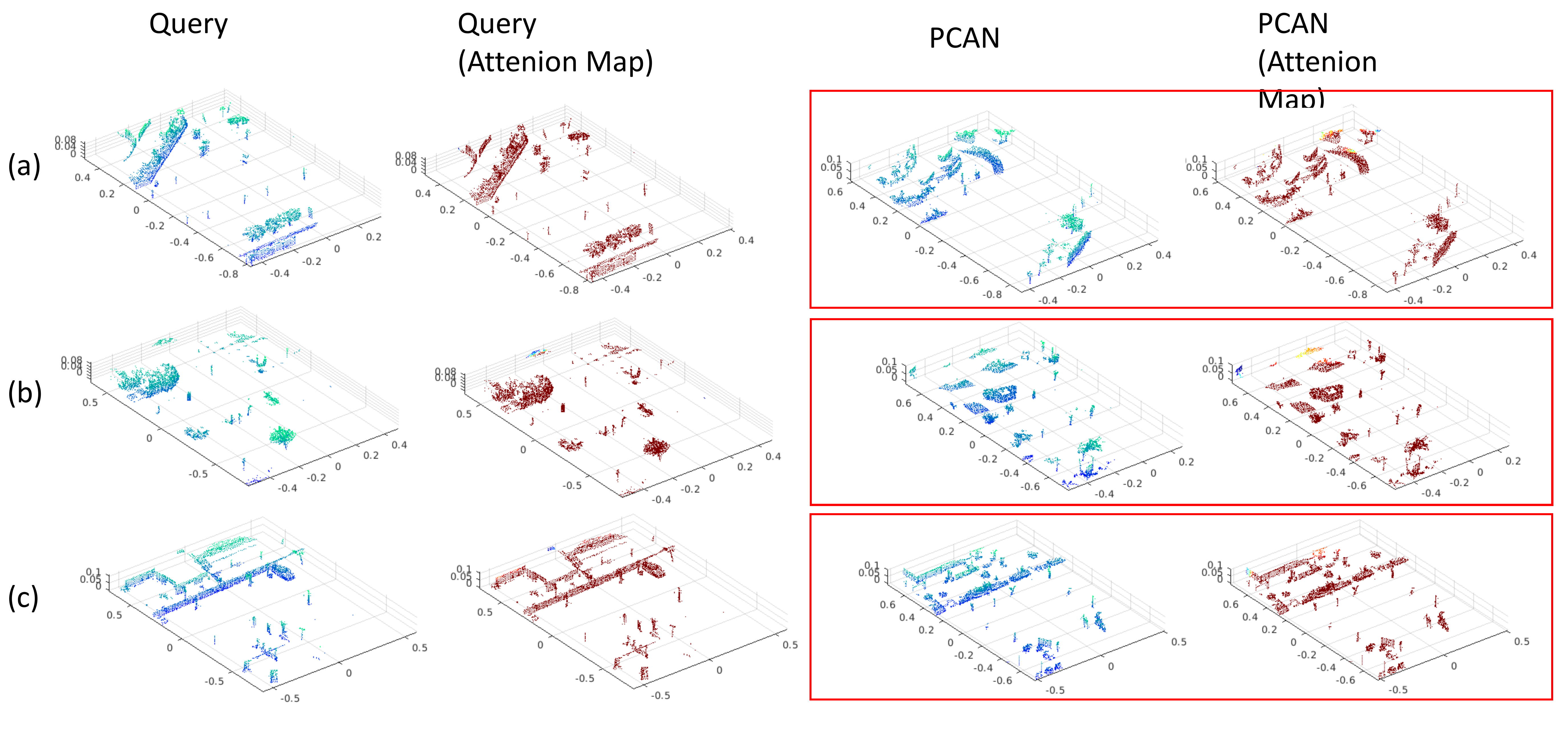}
\caption{Failure cases of our network. It shows the query point cloud with incorrect retrieved results of our network.}
\label{failure}
\end{figure*}

\noindent{\bfseries Refined Network. } Besides training on Oxford datasets, PointNetVLAD does a further comparison with adding U.S. and R.A. to the training sets. We use the same training sets to train a refined network and show the results in Table \ref{refined}. The results show that our network still significantly outperforms the PointNetVLAD on Oxford datasets. It also outperforms our baseline on Oxford dataset. The reason is that adding the U.S. and R.A. training set may have released the overfitting effects on Oxford datasets in some extent.
For in-house datasets, the performance of our refined network is similar to PointNetVLAD. Fig. \ref{recall_carve} shows the recall carves of PCAN and PointNetVLAD for top 25 retrieval results. We show some visualized results in Sec. ~\ref{sec_visualization} and discuss the reason why our refined network does not work effectively on in-house datasets in Sec. ~\ref{discussion}.
\subsection{ Results Visualization}\label{sec_visualization}
We show some rough cases where our refined network can retrieve the correct match while PointNetVLAD can not on Oxford dataset in Fig. \ref{compare}. Each row represents the query point clouds and the retrieval results. First column shows four different query point clouds, while the third and the fifth column show the point clouds retrieved by our network and PointNetVLAD, respectively. The second and the fourth column show the attention maps  obtained during the inference process of our PCAN. The color represents the attention score of each point. The value represented by different color can be referred to the color map at the bottom of Fig. \ref{compare}.

We observe that our network gives low attention scores to the points in a scattered structure which are tree leaves besides the street, while it gives high scores to the buildings (Fig. \ref{compare}(a)-(b)). It implies focusing more on the buildings instead of the trees will reduce the final retrieval losses. It is reasonable since the structure of trees is very unstable due to the gap between tree leaves, resulting that the acquired 3D structure of trees may vary greatly with even a very little movement of the 3D sensors. PointNetVLAD retrieved false point clouds which have a scattered structure of tree leaves while the shape of the buildings is different from the query point cloud.

In Fig. \ref{compare}(c), our network pays little attention to the floating strips and focuses more on the buildings instead. The structures of the strips vary greatly from the query point cloud (row 1) and the correct match (row 3) retrieved by our network shows the unstableness of the strips. PointNetVLAD retrieves a false point cloud with similar strips while the appearance of the buildings varies greatly from the query point cloud.
In Fig. \ref{compare}(d), our network mostly focuses on the corner structure and gives comparatively low scores to the building across the street. This implies the corner structure is more discriminative than planar buildings in retrieval task. The retrieved results of PointNetVLAD only focus on the planar building while ignore the corner structure.

We also show more results of our PCAN in Fig. \ref{good}. These results are from in-house datasets. Though we only train our network on Oxford dataset, it also shows the ability to remove the noisy parts in in-house datasets. From Fig. \ref{good}(a) and (b), our network tries to ignore the parts which are far away from the streets and floats off the grounds.  From Fig. \ref{good}(c), we can see that our network attributes low scores to a bar like object lying in the middle of the street.

\noindent{\bfseries Failure Cases. } Fig. \ref{failure} shows some failure cases of our network. In all the failure cases, our attention network tends to treat all things equally and assigns the same score 1 to all the points in the scene. So in this case, our network degrades to PointNetVLAD.

\subsection{Discussion}\label{discussion}
\noindent{\bfseries Quantity results analysis. } 
We suspect that our refined network dose not work on in-house datasets mainly due to two reasons.
One is the different structures of the point clouds in each database. In Oxford datasets, as the point clouds are almost continuous buildings besides the roads as shown in Fig. \ref{compare}. However, in the in-house datasets, the objects in the scene are more complicated than in Oxford dataset which can be seen in Fig. \ref{good}.
Another reason is that there is insufficient training data of in-house datasets. The number of the point clouds in training sets of oxford dataset are 21711, while it is 6671 of in-house datasets. With more complicated scene structures and less training data of in-house datasets, this makes it harder to learn a network which can produce accurate attention maps for all datasets and our network learns to concentrate more on Oxford dataset than in-house datasets to reach the global minima.

As mentioned in Sec ~\ref{sec_visualization}, in some failure cases, all the outputs of the attention map of in-house datasets tend to all be 1. However, from Fig. \ref{good}, we can see our network still show the ability to ignore the noisy points in the point cloud which is the reason why our baseline network can outperform PointNetVLAD with a small margin in Table \ref{baseline}.
\\
\begin{table}[t]
\centering
\begin{tabular}{|c|c|c|c|c|}
\hline
&\multicolumn{2}{|c|}{Ave recall @1\%} &\multicolumn{2}{|c|}{Ave recall @1\%} \\
\hline
&PCAN &PAN& PCAN &PAN \\
\hline
Oxford& \textbf{83.81} & 81.90& \textbf{69.05} & 65.37 \\
\hline
U.S.& \textbf{79.05} & 78.47& 62.49 &\textbf{62.61} \\
\hline
R.A.& \textbf{71.18} & 69.69& \textbf{57.00} & 54.77 \\
\hline
B.D.& \textbf{66.82} & 63.87& \textbf{58.14} & 57.21\\
\hline
\end{tabular}
\caption{Results show the the average recall (\%) at top 1\% and at top 1 after training on Oxford for PCAN and PAN.}
\label{table_contextual}
\end{table}

\begin{table}[t]
\centering
\begin{tabular}{|c|c|c|c|c|}
\hline
& MRG &MSG\_1 &MSG\_2&MSG\_3 \\
\hline
Oxford& 81.35 &81.56 & 82.10 & \textbf{83.81} \\
\hline
U.S.& 78.72 &79.63 & \textbf{81.62}& 79.05  \\
\hline
R.A.& 66.19 &70.10 &69.16 & \textbf{71.18}\\
\hline
B.D.& 66.75 &66.57 &66.67 & \textbf{66.82}\\
\hline
\end{tabular}
\caption{Results show the the average recall (\%) at top 1\% using different grouping methods.}
\label{grouping}
\end{table}

\noindent{\bfseries Trade-off between stability and discriminating ability. } Based on the analysis of the baseline network, it shows our network has the ability to do a trade off between discriminating ability and stability. On one hand, PCAN gives low scores to the unstable objects such as tree leaves and noisy floating pieces. Though sometimes these unstable objects can be a discriminative marker, our network learns to pay more attention to the buildings which are more robust to the movement of 3D scanners to reduce the global losses. On the other hand, our network can also find the discriminative parts in the scene. Since some buildings are very similar which leads to the lack of the discriminative information, our network learns to find the discriminative part like the corner structure which is a remarkable information for representing a scene.




\subsection{The efficiency of context information}
To evaluate the contribution of the context information in our attention network, we propose another attention network which does not aggregate the contextual information. The attention network uses a PointNet alike architecture which is originally used for semantic segmentation, and we denote it as Point Attention Network(PAN). Refer to the supplementary for details of PAN. 

It can be seen from Table \ref{table_contextual} that our PCAN outperforms PAN on almost all the datasets. Our method outperforms the PAN on Oxford dataset with a margin of 1.9\% at top 1\% retrieved results and 3.6\% at top 1 retrieved results. PCAN also has a better or on-par performance on in-house datasets compared with PAN. It proves that the contextual information have contributions in seeking the task-relevent features in our network.

\subsection{The influences of grouping methodology}
We test the grouping methodology MSG with different scales (1 to 3) and MRG (Multi-resolution grouping)\cite{qi2017pointnet++} with results shown in Table \ref{grouping}. Note that MSG with scale which is bigger than 3 will cause GPU memory problem during training (with GTX 1080Ti 11GB) unless we set a smaller per-point feature size in grouping layer.

\section{Conclusion}
We have proposed a Point Contextual Attention Network which learns point cloud representations using a context-aware features based attention map. Experiments show our PCAN improves the existing state-of-the-art accuracy for point cloud based place recognition. We use only localization tags to get the attention map without additional supervision. The visualization results show the efficiency of our network to predict the significance of the space regions. Due to the flexibility of our network, our introduced PCAN can be easily integrated to other architecture for more point cloud based tasks such as object recognition.

\noindent{\bfseries Acknowledgments} This work was partly supported by The National Key Research and Development Program of China (2017YFB1002600), the NSFC (No. 61672390), Wuhan Science and Technology Plan Project (No. 2017010201010109), and Key Technological Innovation Projects of Hubei Province (2018AAA062). Chunxia Xiao is the corresponding author.

{\small
\bibliographystyle{ieee_fullname}
\bibliography{egbib}
}

\end{document}